\documentclass{article}

\usepackage{arxiv}
\usepackage[utf8]{inputenc}
\usepackage[T1]{fontenc}
\usepackage{hyperref}
\usepackage{url}
\usepackage{booktabs}
\usepackage{amsfonts}
\usepackage{nicefrac}
\usepackage{microtype}
\usepackage[square,sort,comma,numbers]{natbib}
\usepackage{doi}
\usepackage[T1]{fontenc}
\usepackage{lmodern}
\usepackage{cite}
\usepackage{todonotes}
\usepackage{multirow}
\usepackage{tabularx}
\usepackage{subfig}
\usepackage{soul}
\usepackage{subcaption}
\usepackage{algpseudocode}
\usepackage{algorithm}
\usepackage{amsmath}
\usepackage[parfill]{parskip}
\usepackage{nicefrac}

\usepackage{authoraftertitle}
\date{April 2, 2025}

\title{\emph{ZClip}: Adaptive Spike Mitigation for LLM Pre-Training}
\author{
  Abhay Kumar, Louis Owen, Nilabhra Roy Chowdhury, Fabian G\"ura \\
  BluOrion \\
  \texttt{\{abhay.kumar, louis.owen, nilabhra.chowdhury,  fabian.guera\}@bluorion.com} \\
}

\hypersetup{
pdftitle={ZClip: Adaptive Spike Mitigation for LLM Pre-Training},
pdfsubject={cs.CL},
pdfauthor={Abhay Kumar, Louis Owen, Nilabhra Roy Chowdhury, Fabian G\"ura},
pdfkeywords={Gradient Clipping, Loss Spikes, LLM, ZClip},
}

\renewcommand{\undertitle}{}

\begin{document}
\maketitle

\begin{abstract}
Training large language models (LLMs) presents numerous challenges, including gradient instability and loss spikes. 
These phenomena can lead to catastrophic divergence, requiring costly checkpoint restoration and data batch skipping. 
Traditional gradient clipping techniques, such as constant or norm-based methods, fail to address these issues effectively due to their reliance on fixed thresholds or heuristics, leading to inefficient learning and requiring frequent manual intervention. 
In this work, we propose ZClip, an adaptive gradient clipping algorithm that dynamically adjusts the clipping threshold based on statistical properties of gradient norms over time. 
Unlike prior reactive strategies, ZClip proactively adapts to training dynamics without making any prior assumptions on the scale and the temporal evolution of gradient norms.
At its core, it leverages z-score-based anomaly detection to identify and mitigate large gradient spikes, preventing malignant loss spikes while not interfering with convergence otherwise. 
Our code is available at: \url{https://github.com/bluorion-com/ZClip}.
\end{abstract}
\section{Introduction}
\label{sec:introduction}

\begin{figure}[ht]
    \centering
    \includegraphics[width=0.8\textwidth]{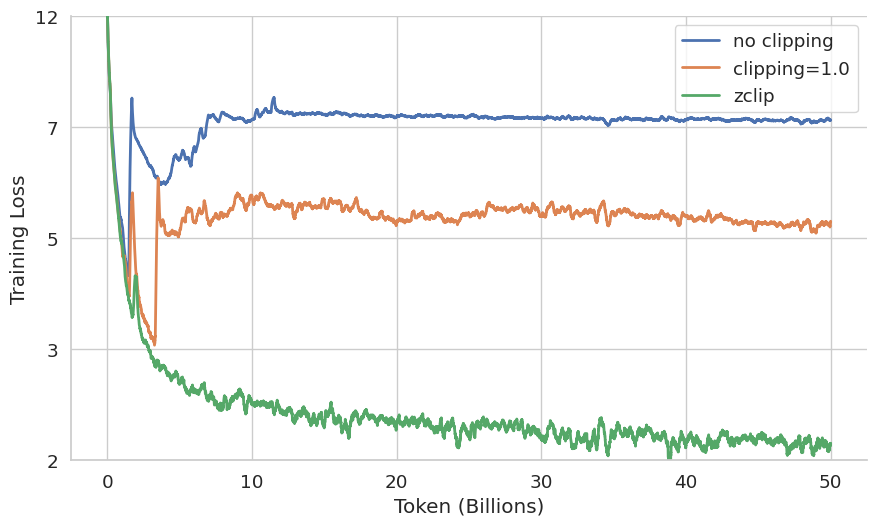}
    \vspace{1em}
\caption{\textbf{Training loss graph comparing 1) training without clipping, 2) clipping with fixed threshold 1.0, and 3) ZClip for a LLaMA 1B model.} 
The learning rate for all three experiments is $3.0 \times 10^{-3}$. 
While both ``no clipping'' and ``constant clipping'' exhibit spiky behavior and diverge early, ZClip (with $z_{\text{thres}}$=2.5 and $\alpha=0.97$) remains stable and continues to optimize effectively throughout training. 
Details on the model configuration and other training hyperparameters are presented in Appendix Section~\ref{sec:llama_config}.}
    \label{fig:lspike_ill}
\end{figure}

Large language models (LLMs) have revolutionized the field of natural language processing \citep{brown2020language, touvron2023llama, deepseekai2024deepseekv2strongeconomicalefficient}. 
However, the training of these models presents significant challenges, among the most critical ones is the occurrence of loss spikes--sudden, extreme increases in the training loss. 
These spikes do not just degrade model performance temporarily, but can also lead to catastrophic divergence, forcing manual intervention to resume training from earlier checkpoints~\citep{geminiteam2024geminifamilyhighlycapable, malartic2024falcon211btechnicalreport}. 

\citep{chowdhery2022palm}, for example, highlights the severity of instability during large-scale language model pre-training. 
In their 540B parameter model training run, the authors observed more than 20 loss spikes throughout the training, each requiring checkpoint rewinds and skipping several hundred batches. 
Interestingly, they note that replaying the same batch that triggered a spike did not consistently reproduce the issue. 
This led them to hypothesize that loss spikes emerge from rare interactions between the optimizer state and specific input batches—a fragile and hard-to-predict combination. 

\citep{liu2025llm360}, as another example, highlights how spikes also increase the environmental impact of LLM pre-training: An additional 30 days were required to address loss spikes while training their 65B model, resulting in the consumption of 129.3 MWh of additional energy. 
To maintain stability, the authors reported resorting to checkpoint rewinds, batch skipping, and learning rate adjustments—manual interventions that increase engineering complexity and compute overhead. 
Notably, they distinguish between two types of loss spikes: \textit{benign} spikes, from which training can recover naturally, and \textit{malignant} spikes, which lead to irreversible divergence. 

Theoretical insights into loss spikes were presented in \citep{takase2024spikemorestabilizingpretraining}, which attributed the issue to the sudden growth of gradient norms. 
By analyzing the spectral norms of Jacobian matrices in Transformer sub-layers, the authors proposed stabilizing training through carefully scaled embeddings and small initialization values. 
While their findings contribute to an early theoretical understanding of loss spikes, the suggested methods apply only statically at initialization time, i.e. once training has started, they can no longer influence training dynamics.

\begin{figure}[ht]
    \centering
    \includegraphics[width=0.8\textwidth]{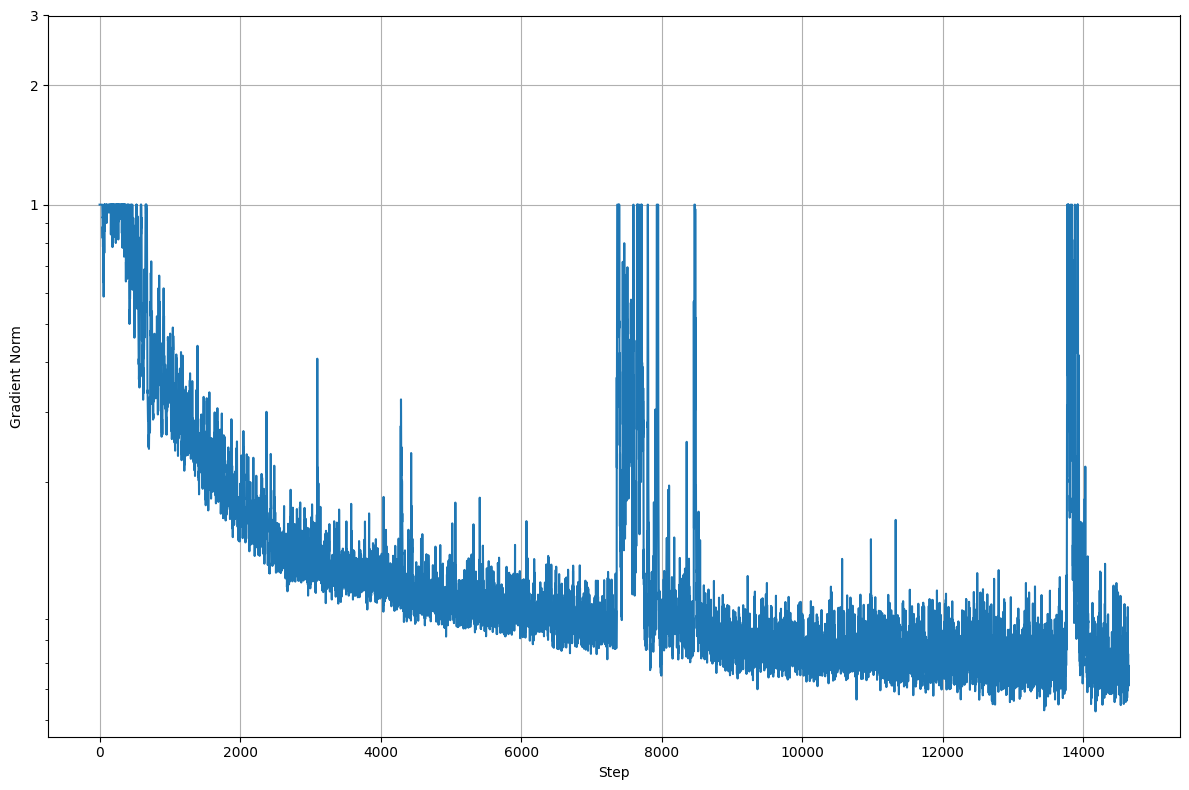} %
    \vspace{1em} %
    \caption{
    \textbf{Training loss graph for a LLaMA 1B model trained with fixed-threshold clipping c = 1.0}.
    Gradient norm spikes persist due to a mismatch between the static threshold and the running distribution. 
    This reveals a key limitation of fixed-threshold clipping in dynamically changing training regimes.}
    \label{fig:constant}
\end{figure}

Gradient clipping has historically been used to address the issue of exploding gradients, particularly in the context of recurrent neural networks (RNNs), where long-term dependencies exacerbate gradient instability~\citep{pascanu2013difficultytrainingrecurrentneural}. 
Traditional gradient clipping methods—used in seminal works such as the original Transformer paper~\citep{vaswani2017attention}, PaLM~\citep{chowdhery2022palm}, and LLaMA~\citep{touvron2023llama, grattafiori2024llama}—typically apply fixed thresholds to control gradient norms. 
While effective in limiting gradient explosion, these approaches lack adaptability to evolving training dynamics and may under-clip in later stages of training (see, for example, Figure~\ref{fig:constant}).

To address these limitations, we propose \textbf{ZClip}, an adaptive gradient clipping algorithm designed to dynamically regulate the gradient norm using recent statistics. 
ZClip employs z-score-based anomaly detection and leverages exponential moving averages (EMA) for robust gradient norm statistics estimation, enabling it to effectively mitigate loss spikes without manual intervention. 
Unlike fixed-threshold gradient norm clipping, ZClip hence adapts to evolving training dynamics.
Through empirical evaluations, we demonstrate that ZClip eliminates loss spikes in all but the most extreme training regimes we tested, and can enable LLMs to achieve baseline performance with a significantly smaller computational and token budget.

\section{Gradient Clipping Methods}

\subsection{Gradient Clipping with Fixed Threshold}

The core idea of gradient clipping is to limit the magnitude of gradients during backpropagation to ensure numerical stability and prevent gradients from becoming excessively large. 
The most common form is fixed-threshold norm clipping, where the gradient norm is constrained to a predefined threshold. 
Mathematically, let $\mathbf{g}_t$ be the gradient vector of a loss function $\mathcal{L}_{\text{data}}(\boldsymbol{\theta}_t)$ with respect to the model parameters $\boldsymbol{\theta}_t$ at training step $t$. 
Then the total gradient norm is defined as:

\begin{equation}
\|\mathbf{g}_t\|_2 = \sqrt{\sum_{i=1}^{N} g_{ti}^2},
\end{equation}

where $N$ is the total number of model parameters. 
If the gradient norm $\|\mathbf{g}_t\|_2$ exceeds a constant threshold $c$, the gradient is scaled down proportionally:

\begin{equation}
\mathbf{g}_t^* =
\begin{cases}
\mathbf{g_t}, & \text{if } \| \mathbf{g}_t \|_2 \leq c, \\
\frac{\mathbf{g}_t}{\| \mathbf{g}_t \|_2} \cdot c, & \text{if } \| \mathbf{g}_t \|_2 > c.
\end{cases}
\end{equation}

This ensures that the updated gradients remain within a controlled range, thus avoiding large parameter updates that could destabilize training. 
While fixed-threshold norm-based gradient clipping has proven effective in traditional deep learning tasks, it exhibits several limitations when applied to modern LLMs.

\paragraph{Inflexibility Under Dynamic Training Regimes.}
One major limitation is the use of a \emph{static} threshold. 
In fixed-threshold gradient clipping, the threshold $c$ remains constant throughout training, despite the general observation that gradient magnitudes vary significantly across iterations and training stages (see, for example, Figure~\ref{fig:constant}). 
The distribution of gradient norms evolves over time—often decreasing in central tendency as the model converges. 
This makes it difficult to select a single, optimal threshold upfront. 
Furthermore, modern LLMs are trained using varying learning rates, curriculum schedules, model depths, batch sizes, and data mixtures. 
These factors induce non-stationary gradient behavior. 
A static threshold cannot in itself account for such evolving dynamics, leading to either unnecessarily slow or unstable training \citep{grattafiori2024llama, deepseekai2024deepseekv2strongeconomicalefficient, liu2025regmixdatamixtureregression, almazrouei2023smolllm}.

\paragraph{Illustrative Example.}
Consider an experiment setup in which the maximum gradient norm is clipped to the threshold $c = 1.0$. 
Let us assume that during the early stages of training, the distribution of gradient norms has a mean value of $\mu \approx 0.8$. 
At a time step $t$, if the gradient norm reaches $\|\mathbf{g}_t\|_2 = 1.2$, the gradient is clipped to $1.0$ as $\|\mathbf{g}_t\|_2 > c$. 
This ensures that the gradient does not deviate too far from the mean $\mu$. Now, as training progresses, the magnitude of the loss decreases (and potentially also the learning rate). 
This will cause the mean to drift towards a lower value, for example, $\mu = 0.2$. 
At this stage of training, if $\|\mathbf{g}_t\|_2 = 0.9$ is encountered, no clipping will take place as $\|\mathbf{g}_t\|_2 < c$. 
However, this can be detrimental for training as $\|\mathbf{g}_t\|_2$ is much higher than the current mean $\mu$ and the high gradient norm can adversely affect the model weights.

Such a deviation suggests that the distribution's tail is being reached or even surpassed more frequently, potentially leading to instabilities (see, for example, Figure~\ref{fig:constant}). 
This example highlights the limitation of fixed-threshold clipping: it fails to adjust for shifts in the underlying gradient norm distribution over time.

\paragraph{Sensitivity to Hyperparameters.}
The optimal threshold $c$ furthermore depends on the particular model and the many hyperparameters in the training recipe. 
If $c$ is too small, gradients are over-clipped and learning slows down. If $c$ is too large, the gradients may not be clipped sufficiently, resulting in instability. 
All of these dependencies make tuning non-trivial, especially for large-scale models.

In summary, fixed-threshold clipping suffers from two key limitations: (a) the threshold $c$ does not in itself evolve with training, and (b) it is sensitive to model-specific factors. 
These shortcomings motivate the need for adaptive methods, as discussed in the following.

\subsection{AutoClip}
AutoClip~\citep{seetharaman2020autoclipadaptivegradientclipping} computes an adaptive clipping threshold by selecting a user-defined percentile from the (pre-clipping) gradient norm history. 
It was originally proposed in the context of an audio source separation task using a bi-directional LSTM model.
While applicable in principle in the context of LLMs, storing every gradient norm over millions of steps incurs some memory and compute overhead.
Furthermore, the use of the pre-clipping gradient norm history still makes it sensitive to outliers accumulating over the course of long LLM pre-training runs.

\subsection{ZClip}
To address the aforementioned challenges related to the temporal evolution of gradient norms during training, model-specific hyperparameters, stability in extremely long training regimes, and the memory and compute overhead of maintaining a full gradient norm history, we propose ZClip. 
At its core, the approach relies on the z-score statistics of the gradient norm to detect abnormal spikes (see Section~\ref{detection}), and to adjust the clipping dynamically (see Section~\ref{adjustment}). 
This allows ZClip to respond effectively to the evolving training dynamics. 
Since we are using z-scores, we assume that gradient norms over a short window are approximately normally distributed (see Appendix~\ref{appendix:normality} for an empirical analysis).
To capture this local behavior of the gradient norms, ZClip tracks the exponential moving average (EMA) of both mean and variance of the gradient norm. 
In contrast to methods such as AutoClip, ZClip only maintains a lightweight summary of the recent history. As a result, it is more memory- and compute-efficient, making it particularly suitable for large-scale training scenarios.

\subsubsection{Z-score based Spike Detection}
\label{detection}
The ZClip process begins by computing the norm of the gradient $\mathbf{g}_t$ at training step $t$ as:
\begin{equation}
g_t = \| \mathbf{g}_t \|_2.
\end{equation}

ZClip then iteratively updates the mean and standard deviation with exponential smoothing. 
Specifically, the mean $\mu_t$ and the standard deviation $\sigma_t$ are updated with a smoothing factor $\alpha \in (0, 1)$ such that:
\begin{equation}
\mu_t = \alpha \mu_{t-1} + (1 - \alpha) g_t,
\end{equation}
\begin{equation}
\sigma_t = \sqrt{\alpha \sigma_{t-1}^2 + (1 - \alpha)\,\bigl(g_t - \mu_t\bigr)^2}.
\end{equation}
We then evaluate the deviation of the current gradient norm from the running mean in terms of standard deviations. This is achieved through the z-score calculation:
\begin{equation}
z_t = \frac{(g_t - \mu_t)}{\sigma_t}.
\end{equation}
The z-score $z_t$ then represents how many standard deviations the current gradient norm lies away from the mean.
A spike is detected if $z_t > z_{\text{thres}}$.

\subsubsection{Z-score based Gradient Adjustment}
\label{adjustment}
If, based on the z-score, a spike is detected, we once again rely on the z-score to adjust the gradient norm.
We define the adjusted gradient as:
\begin{equation}
\mathbf{g}_t^* =
\begin{cases}
\mathbf{g}_t & \text{if } z_t \leq z_{\text{thres}} \\
\frac{\mathbf{g}_t}{\| \mathbf{g}_t \|_2} (\mu_t + z_t^* \cdot \sigma_t ) & \text{if } z_t > z_{\text{thres}},
\end{cases}
\end{equation}
where
\begin{equation}
    z_t^* = \xi(z_t(g_t)).
\end{equation}
The function $\xi(z_t)$ maps the original z-score $z_t$ to an adjusted, ``clipped'' z-score $z_t^*$.
To reduce the gradient norm for spikes, $\xi$ must satisfy $\xi(z_t) \leq z_t $ for $z_t \geq z_{\text{thres}}$.
If $\xi(z_t) \leq z_{\text{thres}}$ for $z_t \geq z_{\text{thres}}$, then the gradient norm is strictly clipped to $g_t^* \leq \mu_t + z_{\text{thres}} \cdot \sigma_t$.
For continuity at $z_t = z_{\text{thres}}$, one may want to impose $\xi(z_t = z_{\text{thres}}) = z_{\text{thres}}$ in addition.

\begin{figure}[ht]
    \centering
    \includegraphics[width=0.8\textwidth]{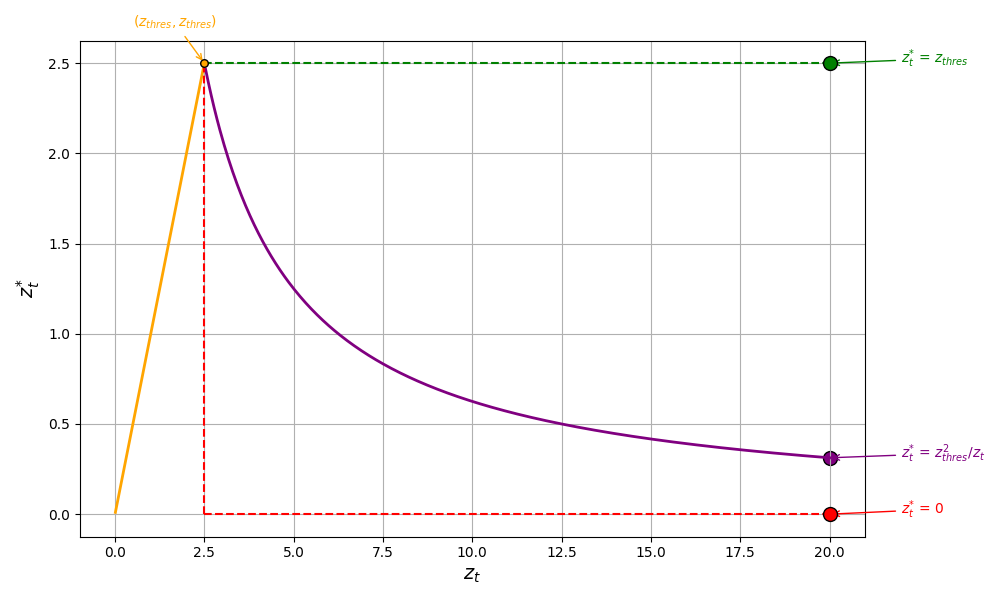} 
    \vspace{1em} %
    \caption{
    \textbf{Three possible choices for the z-score adjustment function $\xi(z_t)$ illustrated for $z_{\text{thres}} = 2.5$.}
    Note the discontinuity for $\xi(z_t) = 0$, and the reciprocal nature of $\xi(z_t) = \nicefrac{z_{\text{thres}}^2}{z_t}$ leading to more aggressive clipping for more extreme outliers.}
    \label{fig:zt_zstar_2_5}
\end{figure}

Figure ~\ref{fig:zt_zstar_2_5} illustrates three possible choices for $\xi$:
\begin{itemize}
  \item \textbf{Clipping to mean}: $z_t^* = 0$ (red line) clips the gradient norm to the mean $\mu_t$, creating a discontinuity at $z_t = z_{\text{thres}}$.
  \item \textbf{Clipping to max}: $z_t^* = z_{\text{thres}}$ (green line) clips the gradient norm to the maximum value $g_t^* = \mu_t + z_{\text{thres}} \sigma_t$. 
  It is continuous at $z_t = z_{\text{thres}}$.
  \item \textbf{Reciprocal clipping}: $z_t^* = \nicefrac{z_{\text{thres}}^2}{z_t}$ (purple line) is a possible compromise between the two previous strategies. 
  It is continuous at $z_t = z_{\text{thres}}$, but clips extreme outliers more aggressively and eventually converges to ``clip to mean'', i.e. $\lim_{z_t \to \infty} g_t^* = \mu_t$.
\end{itemize}

We experimented with all three options presented for $\xi(z_t)$ and observed the best results for ``reciprocal clipping'' (see Appendix~\ref{sec:clip-to-mean-max}).
In the following, we therefore assume $\xi(z_t) = \nicefrac{z_{\text{thres}}^2}{z_t}$ unless otherwise specified.
We hypothesize that adjusting the clipping strength dynamically based on the severity of the anomaly helps mitigate the potentially destabilizing effect of discontinuous gradient norm adjustments, as well as the algorithm's sensitivity towards a particular (potentially slightly suboptimal) choice of $z_{\text{thres}}$.
If one interprets $\eta = \nicefrac{z_t}{z_{\text{thres}}}$ as a signal-to-noise ratio (SNR), then scaling by the severity $\eta^{-1}$ can intuitively be interpreted as follows:
When the gradient norm is very high, it generally indicates the presence of significant noise, thus requiring stronger clipping to filter out this disruptive component; in contrast, when the gradient norm is only slightly elevated, it implies lower noise levels, so less aggressive clipping suffices to preserve the informative signal.

\subsubsection{Initialization via Warm-up} 
The initialization of the mean and variance estimates plays a critical role in the performance of ZClip. We experimented with several schemes, including fixed constants, early exponential moving average (EMA) bootstrapping, and dynamic ramp-up strategies. However, we found that using a short warm-up period yielded the most stable results.

During the first $N_w$ steps (e.g., $N_w=25$), we collect the unmodified gradient norms without applying ZClip. At the end of this warm-up phase, we compute the initial mean and variance as follows:
\begin{equation}
\mu_{N_w} = \frac{1}{N_w} \sum_{t=1}^{N_w} g_t, \quad 
\sigma_{N_w} = \sqrt{\frac{1}{N_w} \sum_{t=1}^{N_w} (g_t - \mu_{N_w})^2}.
\end{equation}
These values serve as the starting point for the EMA-based updates that continue throughout the remainder of the training. 
This approach ensures that the statistics are grounded in the actual behavior of the gradients at the beginning of training, providing a stable foundation for subsequent anomaly detection and clipping.

\subsubsection{Handling Spikes in Statistics Updates} 
A key challenge in maintaining accurate statistics arises when the current gradient norm $g_t$ is classified as a spike, that is,  $z_t > z_\text{thres}$. 
Including such extreme values in the updates can skew the moving averages and inflate future thresholds, leading to reduced sensitivity towards anomalies. 
On the other hand, skipping the update entirely biases the statistics toward lower values, which can make clipping overly aggressive over time.

To strike a balance, we use the clipped gradient norm $g_t^*$ to update the statistics during such events. 
Specifically, we define the value used to update the statistics as:
\begin{equation}
g_t^\text{update} =
\begin{cases}
g_t, & \text{if } z_t \leq z_\text{thres}, \\
g_t^* & \text{if } z_t > z_\text{thres}.
\end{cases}
\end{equation}
This update strategy ensures that the statistics remain representative of the stable training regime while still adapting to changing dynamics. 

\subsubsection{Algorithm \& Implementation} 
Algorithm~\ref{alg:ema_clip_revised} summarizes the mathematical formulations presented above and illustrates ZClip from a practical implementation perspective.
It incorporates the key steps outlined above, namely warm-up, anomaly detection, gradient norm adjustment, and the statistics update rule in case of spikes.
A concrete implementation for use with PyTorch (Lightning) can be found at \url{https://github.com/bluorion-com/ZClip}.

\begin{algorithm}
\caption{Training Loop with ZClip (EMA-based updates; Reciprocal clipping)}
\label{alg:ema_clip_revised}
\begin{algorithmic}

\Require
  \Statex \quad $\alpha \in (0,1)$ \Comment{EMA smoothing factor (e.g., 0.97)}
  \State  \quad $z_{\text{thres}} \in \mathbb{R}^+$ \Comment{Z-score threshold (e.g., 2.5)}
  \State  \quad $\epsilon \gets 10^{-6}$ \Comment{Small positive constant}
  \Statex \quad $\mathit{lr} > 0$ \Comment{Learning rate (e.g., 0.001)}
  \Statex \quad $f(\boldsymbol{\theta})$ \Comment{Objective function}
  \Statex \quad $\mu_t, \mathit{v_t} \in \mathbb{R}$ \Comment{EMA estimates of gradient norm mean and variance}
  \Statex \quad $N_w \in \mathbb{N}$ \Comment{Number of warm-up steps (e.g., 25)}
  \Statex \quad $t_{\text{max}} \in \mathbb{N}$ \Comment{Total number of training steps}
  
\State \textbf{Initialize via warm-up:} Collect $N_w$ gradient norms to compute $\mu_{N_w}$ and $\mathit{v_{N_w}}$
\State $t \gets N_w$

\While{$t < t_{\text{max}}$}
  
  \State $\mathbf{g}_t \gets \nabla f_t(\boldsymbol{\theta}_{t})$ \Comment Gradient computation
  \State $g_t \gets \|\,\mathbf{g}_t\,\|_2$
  \State $z_t \gets \dfrac{g_t - \mu_t}{\sqrt{\mathit{v_t}} + \epsilon}$
  \State $g_{t}^* \gets g_t$

  \If{$z_t > z_{thres}$}
    \State $g_{t}^* \gets \mu_t + \dfrac{z_{thres}^2}{\mathit{z_t}} \sqrt{\mathit{v_t}}$
  \EndIf

  \State \text{clip\_grad\_norm\_}($g_t^*$) \Comment{PyTorch in-place gradient clipping}
  \State $\boldsymbol{\theta}_{t+1} \gets \text{optimizer\_update}(\boldsymbol{\theta}_{t}, lr, \mathbf{g}_t)$

  \State $\mu_{t+1} \gets \alpha \mu_t + (1 - \alpha) g_t^*$ \Comment EMA update for mean
  \State $\mathit{v_{t+1}} \gets \alpha \mathit{v_{t}} + (1 - \alpha) \left(g_t^* - \mu_{t+1}\right)^2$ \Comment EMA update for variance

  \State $t \gets t + 1$
\EndWhile
\State \Return $\boldsymbol{\theta}_{t_\text{max}}$

\end{algorithmic}
\end{algorithm}

\section{Experiment Setup}
In the following, we describe the setup of the experiments we ran to answer three core questions:
\begin{enumerate}
    \item Does ZClip effectively mitigate spikes in pre-training, especially in ``aggressive'', spike-prone training regimes such as when running with high learning rates?
    \item If so, does ZClip stabilize these ``aggressive'' regimes enough such that training converges towards a particular loss or benchmark milestone earlier than in ``standard'' training regimes, consequently reducing the total computational cost?
    \item In ``standard'' training regimes, i.e. regimes with learning rates that are widely used in literature and are observed to spike occasionally, are there any negative consequences of activating ZClip in terms of loss convergence, downstream task performance, or throughput?
\end{enumerate}

\subsection{Model Setup}
Unless otherwise specified, all experiments were performed using a 1B (16 layers) LLaMA \citep{touvron2023llamaopenefficientfoundation} model.
The training dataset consisted of the SmolLM \citep{allal2025smollm2smolgoesbig} corpus, which in turn is comprised of three sources: FineWebEdu-Deduplicated, Cosmopedia-V2, and Python-Edu. 
From this corpus we randomly sampled 50 billion tokens (50BT) and employed packing to obtain examples with a fixed context length of 2048.
All experiments were conducted in a distributed training setup using the Fully Sharded Data Parallelism (FSDP) strategy across four GPU nodes, each equipped with eight H100 GPUs. 
Unless explicitly stated otherwise, ZClip was configured with default hyperparameters \(\alpha = 0.97\) for the smoothing factor, and threshold \(z_{\text{thres}} = 2.5\). 
The results of the hyperparameter sweeps used to determine these values are presented in Appendix~\ref{appendix:ablation}.
Other training hyperparameters are presented in Table~\ref{tab:training_hyperparams}.
More details regarding the model configuration and tokenizer are provided in Appendix \ref{Model and Tokenizer Details}.

\begin{table}[H]
\centering
\begin{tabular}{|l|p{9cm}|}
\hline
\textbf{Hyperparameter} & \textbf{Value} \\
\hline
Optimizer & Fused AdamW ($\beta_1 = 0.9$, $\beta_2 = 0.999$, $\epsilon = 1 \times 10^{-7}$) \\
Learning Rate Schedule & Linear warm-up followed by cosine decay \\
Max. Learning Rate & $1 \times 10^{-4}$ to $5 \times 10^{-3}$ \\
End Learning Rate & 10\% of Maximum Learning Rate \\
Warm-up Tokens & 2 billion tokens (2BT) \\
Weight Decay & 0.01 (AdamW implementation) \\
Global Batch Size & 2048 \\
Sequence Length & 2048 \\
Precision & Mixed Precision BFloat16 \\
\hline
\end{tabular}
\vspace{10pt}
\caption{\textbf{Default training hyperparameters for the presented 1B LLaMA pre-training experiments.}}
\label{tab:training_hyperparams}
\end{table}

\subsection{Evaluation Metrics}
To assess the effectiveness of ZClip, we employed a combination of qualitative and quantitative metrics that capture various aspects of training stability, model quality, and gradient behavior:
\begin{itemize}
    \item Training stability was qualitatively evaluated by examining the smoothness of the loss curve and the number of (or absence of) loss spikes during training. 
    A stable training process should exhibit consistent convergence without abrupt fluctuations, which can disrupt optimization and lead to suboptimal performance. 
    \item The quality of the model was measured by evaluating its performance on downstream benchmark tasks, specifically HellaSwag \citep{zellers2019hellaswagmachinereallyfinish} and WinoGrande \citep{sakaguchi2019winograndeadversarialwinogradschema}. 
    These benchmarks provide insight into how well the model generalizes and performs on real-world tasks, serving as a proxy for the impact of gradient clipping on overall model effectiveness.
    \item To analyze gradient behavior, we tracked in particular the mean and standard deviation of the gradient norm over time.
    This analysis helped us understand the behavior of the clipping mechanism and its ability to adapt to the dynamic nature of training. 
\end{itemize}

\subsection{Learning Rate Regimes}
We evaluated ZClip across the spectrum of feasible learning rates:

\textbf{High learning rates} (in our scenario $5.0 \times 10^{-3}$ - $3.0 \times 10^{-3}$) often lead to unstable training and, if not properly managed, can result in irrecoverable divergence. 
However, training with high learning rates has the potential to enable faster convergence, which can significantly reduce training time and token budget requirements. 

At \textbf{lower learning rates} (in our scenario $1.0 \times 10^{-3}$ - $1.0 \times 10^{-4}$), training becomes more stable, but small gradients may still cause subtle instabilities that accumulate over time. 
This scenario evaluates whether ZClip can effectively handle minor fluctuations without overly regularizing benign updates, ensuring that the method remains adaptive and efficient even in more stable regimes.

\section{Results and Analysis}
\label{sec:results}

\subsection{Performance at High Learning Rates}
\begin{figure}[htbp]
    \centering
    \includegraphics[width=0.7\textwidth]{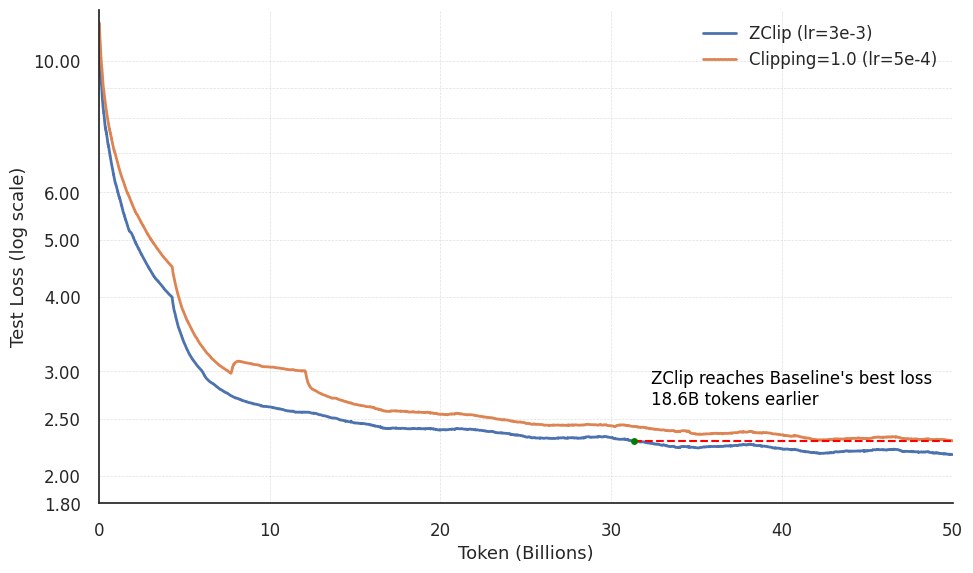} %
    \vspace{1em} %
    \caption{\textbf{Comparison of Test Loss between ZClip (lr=3e-3) and a baseline model (lr=5e-4) on a 50B token corpus}. 
    ZClip achieved the same final loss as the baseline while requiring 18.6B fewer tokens to get there. 
    The plot uses log-scaled training loss for visibility, and smoothing has been applied to reduce noise. 
    ZClip allows for faster convergence without compromising on final loss value.}
    \label{fig:low_high}
\end{figure}

At higher learning rates, such as $3.0 \times 10^{-3}$, enabling ZClip led to noticeable improvements in training stability, loss convergence, and downstream benchmark performance (see Table~\ref{tab:zclip_comparison} and Figure~\ref{fig:low_high}). 
In the particular scenario presented, the model trained with ZClip and a learning rate of $3.0 \times 10^{-3}$ reached the best baseline validation loss more than 35\% faster in terms of training steps compared to a baseline model trained with fixed-threshold gradient norm clipping and the best learning rate of $5.0 \times 10^{-4}$. 
For a learning rate of $3.0 \times 10^{-3}$ the baseline did not converge.
By supporting stable optimization at more aggressive learning rates, ZClip expands the viable hyperparameter space and accelerates convergence. 

\begin{table}[htbp]
    \centering
    \begin{tabular}{|c|cc|cc|cc|}
        \hline
        \textbf{Learning} & \multicolumn{2}{c|}{\textbf{Train Loss}} & \multicolumn{2}{c|}{\textbf{HellaSwag}} & \multicolumn{2}{c|}{\textbf{WinoGrande}} \\
        \cline{2-7}
         \textbf{Rate} & \textbf{Fixed Threshold} & \textbf{ZClip} & \textbf{Fixed Threshold} & \textbf{ZClip} & \textbf{Fixed Threshold} & \textbf{ZClip} \\
         \hline
         5e-3  & 7.28 & \textbf{5.91} & \textbf{25.89} & 25.27 & \textbf{49.32} & 48.46 \\
         3e-3  & 5.14 & \textbf{2.14} & 25.60       & \textbf{50.82} & 49.03       & \textbf{53.27} \\
         1e-3  & 2.34 & \textbf{2.18} & 43.01       & \textbf{49.30} & 52.32       & \textbf{54.85} \\
         7e-4  & 2.29 & \textbf{2.20} & 44.27       & \textbf{46.94} & \textbf{54.14} & 52.72 \\
         5e-4  & 2.27 & \textbf{2.24} & 45.27       & \textbf{45.62} & 52.32       & \textbf{52.48} \\
         3e-4  & 2.34 & \textbf{2.33} & 42.34       & \textbf{42.56} & 50.35       & \textbf{51.53} \\
         1e-4  & 2.46 & 2.46 & 33.92       & \textbf{34.12} & 51.22       & \textbf{53.35} \\
         \hline
    \end{tabular}
    \vspace{10pt}
    \caption{\textbf{Comparison of ZClip to fixed-threshold clipping across a range of learning rates on train loss and downstream task performance.} For learning rates of $3\times10^{-3}$ and $5\times10^{-3}$, we set $z_{\text{thres}}=2.0$. For the other experiments, the clipping parameters were $z_{\text{thres}}=2.5$ for ZClip and $c=1.0$ for fixed-threshold clipping, respectively.}

    \label{tab:zclip_comparison}
\end{table}

\begin{figure}[htbp]
  \centering
  \captionsetup[subfloat]{labelformat=empty} %

  \subfloat{\includegraphics[width=0.3\textwidth]{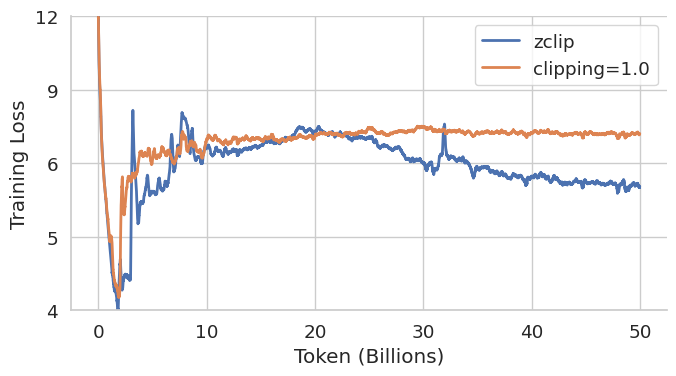}}\hfill
  \subfloat{\includegraphics[width=0.3\textwidth]{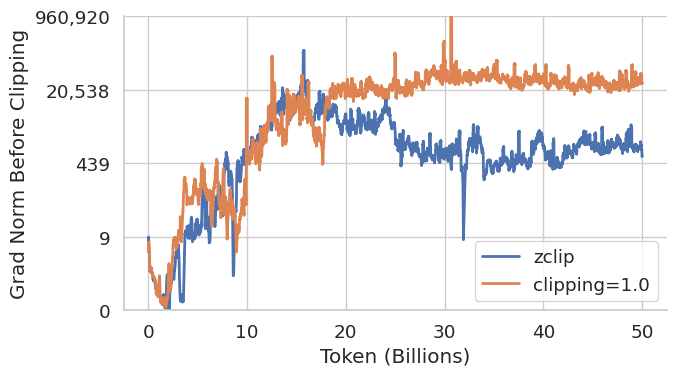}}\hfill
  \subfloat{\includegraphics[width=0.3\textwidth]{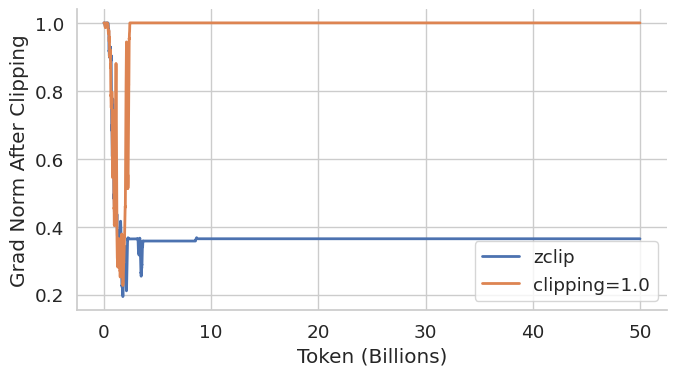}}
  \par\medskip\textbf{Learning Rate = \(5.0 \times 10^{-3}\)}\par\bigskip

  \vspace{1em} %

  \subfloat{\includegraphics[width=0.3\textwidth]{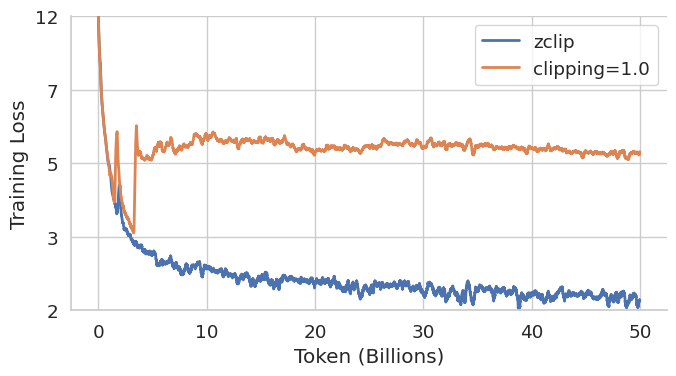}}\hfill
  \subfloat{\includegraphics[width=0.3\textwidth]{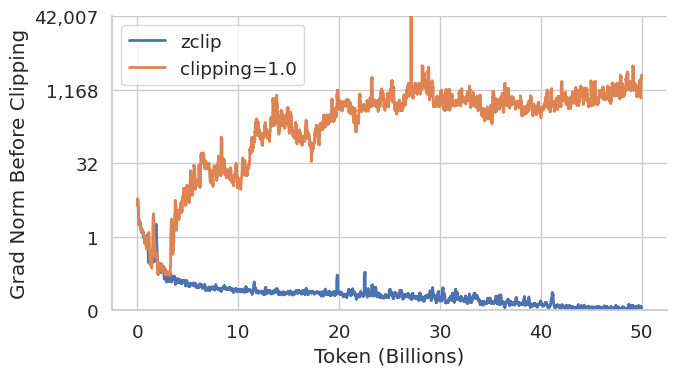}}\hfill
  \subfloat{\includegraphics[width=0.3\textwidth]{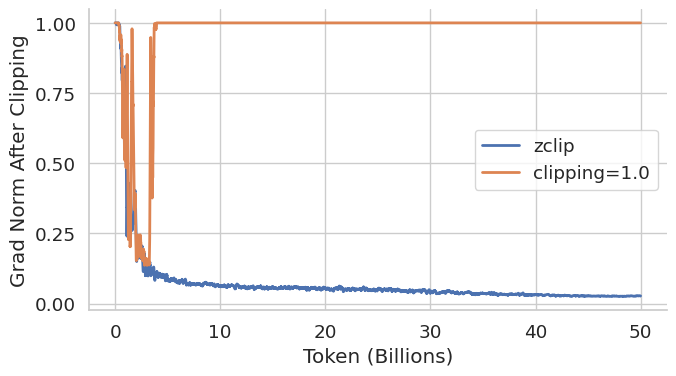}}
  \par\medskip\textbf{Learning Rate = \(3.0 \times 10^{-3}\)}\par\bigskip

  \caption{\textbf{ZClip and fixed-threshold clipping at higher learning rates.} 
  Each row shows training loss (left), gradient norm before clipping (middle), and after clipping (right). 
  For \(3.0 \times 10^{-3}\) ZClip stabilized gradients and reduces post-clipping spikes, unlike fixed-threshold clipping which accumulates instability.
  For \(5.0 \times 10^{-3}\) both clipping methods saturated.
  }
  \label{fig:lr_exp_high}
\end{figure}

That being said, we caution that ZClip is not a complete substitute for principled learning rate tuning—extremely high learning rates can still lead to divergence.
At an even higher learning rate of \(5.0 \times 10^{-3}\), both fixed-threshold clipping and ZClip diverged. 
As shown in Figure~\ref{fig:lr_exp_high}, the gradient norm remained persistently high, causing continuous clipping for both strategies and ultimately leading to divergence. 

\subsection{Performance at Lower Learning Rates}
\begin{figure}[htbp]
  \centering
  \captionsetup[subfloat]{labelformat=empty} %

  \subfloat{\includegraphics[width=0.3\textwidth]{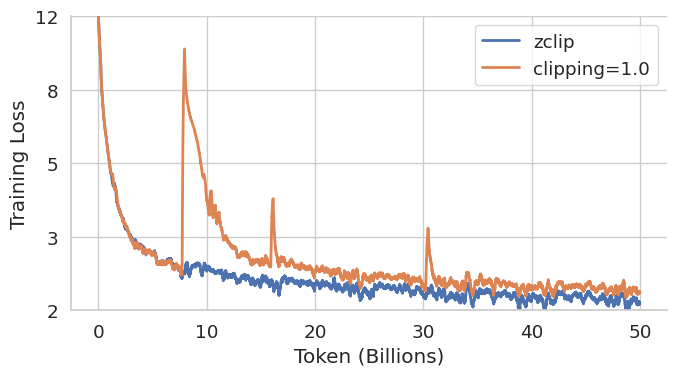}}\hfill
  \subfloat{\includegraphics[width=0.3\textwidth]{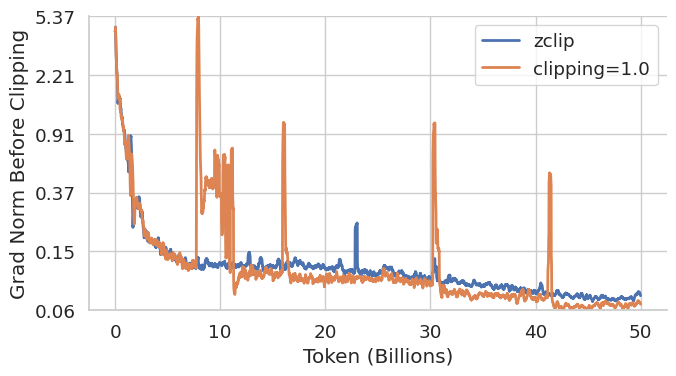}}\hfill
  \subfloat{\includegraphics[width=0.3\textwidth]{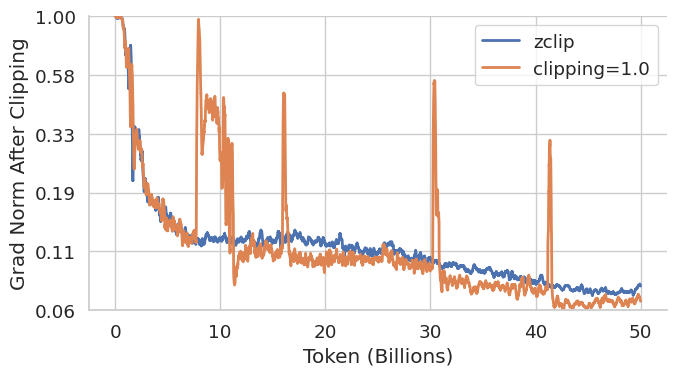}}
  \par\medskip\textbf{Learning Rate = \(1.0 \times 10^{-3}\)}

  \subfloat{\includegraphics[width=0.3\textwidth]{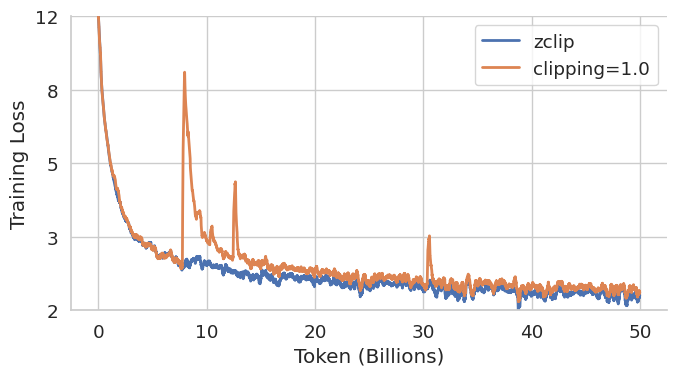}}\hfill
  \subfloat{\includegraphics[width=0.3\textwidth]{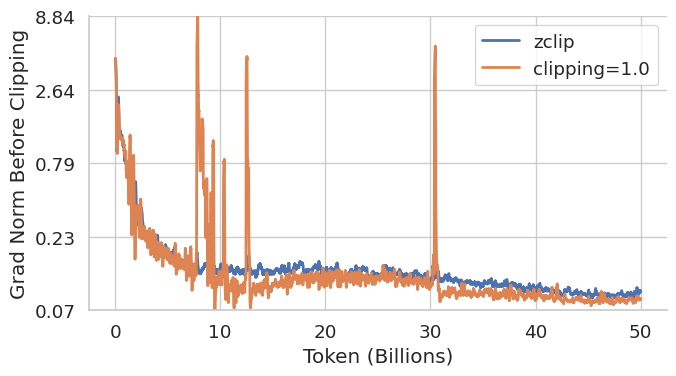}}\hfill
  \subfloat{\includegraphics[width=0.3\textwidth]{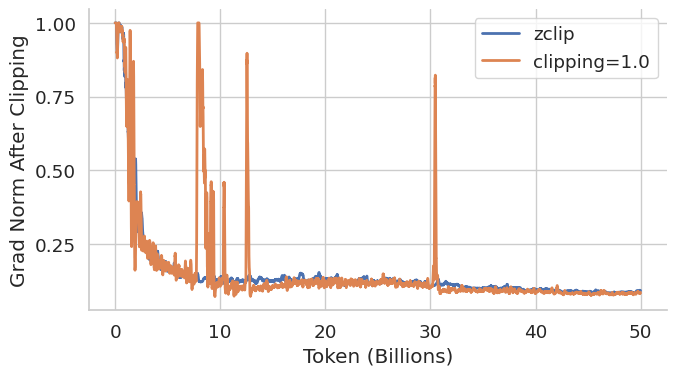}}
  \par\medskip\textbf{Learning Rate = \(7.0 \times 10^{-4}\)}

  \subfloat{\includegraphics[width=0.3\textwidth]{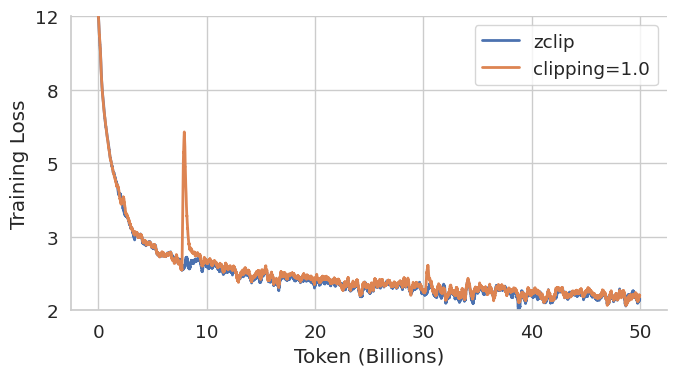}}\hfill
  \subfloat{\includegraphics[width=0.3\textwidth]{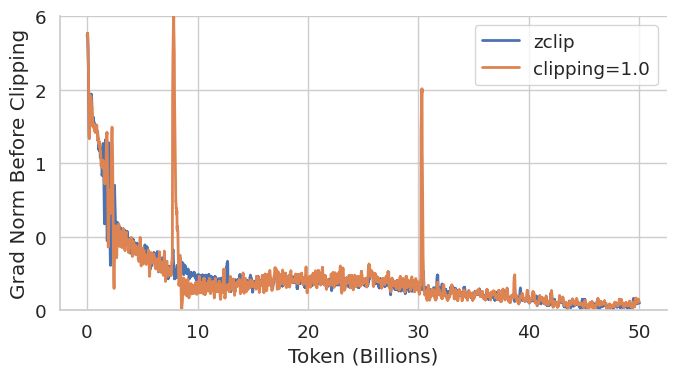}}\hfill
  \subfloat{\includegraphics[width=0.3\textwidth]{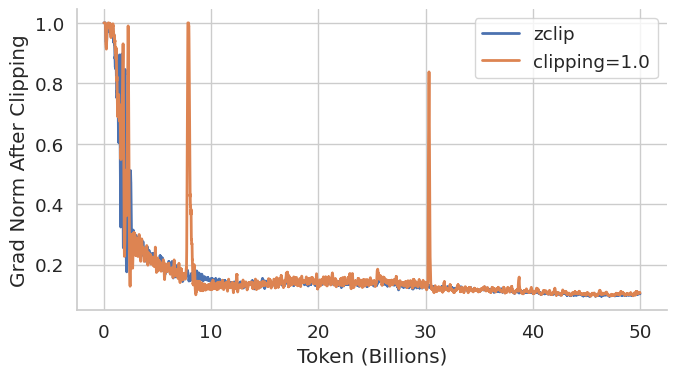}}
  \par\medskip\textbf{Learning Rate = \(5.0 \times 10^{-4}\)}

  \subfloat{\includegraphics[width=0.3\textwidth]{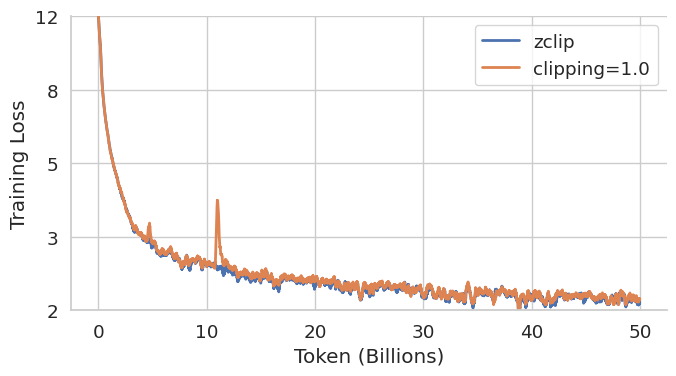}}\hfill
  \subfloat{\includegraphics[width=0.3\textwidth]{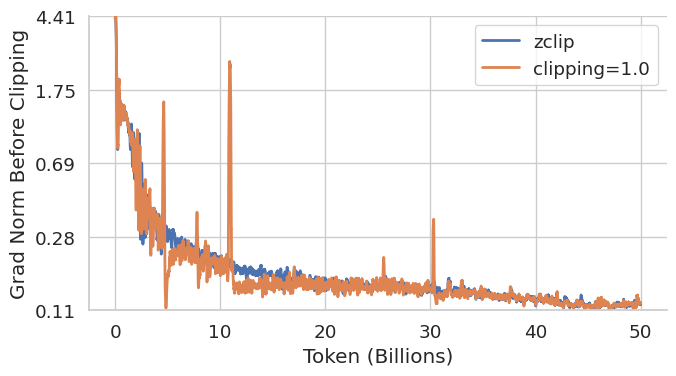}}\hfill
  \subfloat{\includegraphics[width=0.3\textwidth]{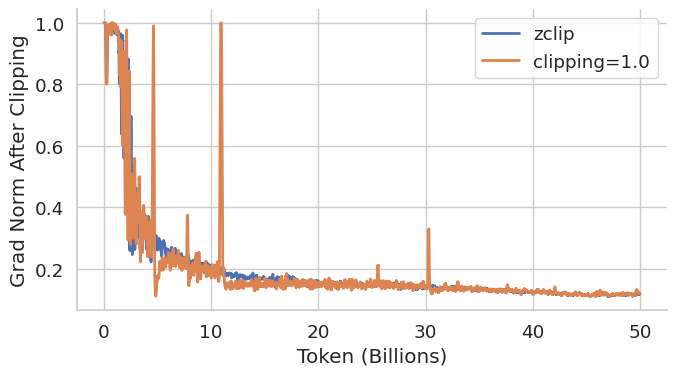}}
  \par\medskip\textbf{Learning Rate = \(3.0 \times 10^{-4}\)}

  \subfloat{\includegraphics[width=0.3\textwidth]{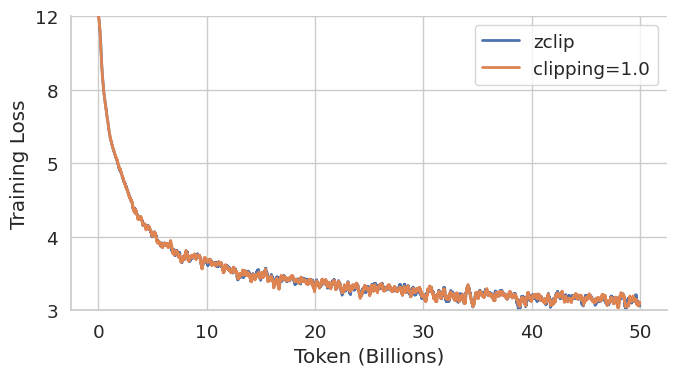}}\hfill
  \subfloat{\includegraphics[width=0.3\textwidth]{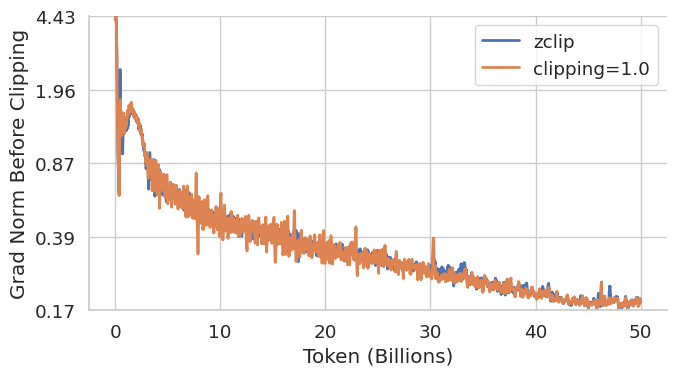}}\hfill
  \subfloat{\includegraphics[width=0.3\textwidth]{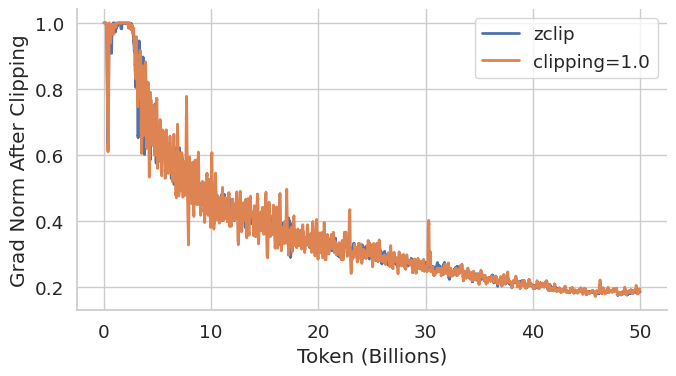}}
  \par\medskip\textbf{Learning Rate = \(1.0 \times 10^{-4}\)}

  \caption{\textbf{ZClip and fixed-threshold clipping at lower learning rates}. Each row shows training loss (left), gradient norm before clipping (middle), and after clipping (right). 
  ZClip preserves stability and convergence also at smaller learning rates, while fixed-threshold still struggles with (benign) spikes.}
  \label{fig:lr_exp_normal}
\end{figure}

At lower learning rates, ZClip demonstrated its ability to handle minor gradient fluctuations without over-regularizing benign updates, resulting in slightly better performance compared to fixed-threshold gradient norm clipping method (see Table~\ref{tab:zclip_comparison} and Figure~\ref{fig:lr_exp_normal}). 
We hypothesize that this is due to the ability of ZClip to suppress high-frequency gradient noise more effectively in advanced training stages.
In the bigger picture, ZClip does not seem to degrade performance in already stable settings, preserving the natural convergence behavior of the model.

\begin{table}%
\centering
\begin{tabular}{|l|c|c|c|c|}
\hline
\textbf{Method} & \textbf{Spike Count} & \textbf{Test Loss} & \textbf{HellaSwag (\%)} & 
\textbf{WinoGrande (\%)} \\
\hline
No Clipping      & 12 & 2.90 & 31.01  & 51.01  \\
Clipping=1.0     & 6 & 2.33  & 43.01  & 52.32 \\
Autoclip     & 0 & 2.20  & 48.10  & 53.67 \\
ZClip   & \textbf{0} & \textbf{2.18} & \textbf{49.30} & \textbf{54.85} \\
\hline
\end{tabular}
\vspace{10pt}
\caption{\textbf{Performance comparison of gradient clipping methods on HellaSwag and Winogrande after 50BT with a learning rate of 1e-3.}}
\label{tab:stability}
\end{table}

In the following, we focus on the learning rate of \(1.0 \times 10^{-3}\), as it strikes a balance between being high enough to challenge stability, and low enough to allow for meaningful gradient dynamics when comparing ZClip against other clipping methods. 
Table~\ref{tab:stability} provides the number of loss spikes observed during training for the four scenarios under investigation; both AutoClip and ZClip managed to suppress all spikes, but ZClip outperformed in terms of downstream benchmarks.

\begin{figure}[htbp]
    \centering
    \subfloat[]{%
        \includegraphics[width=0.49\textwidth]{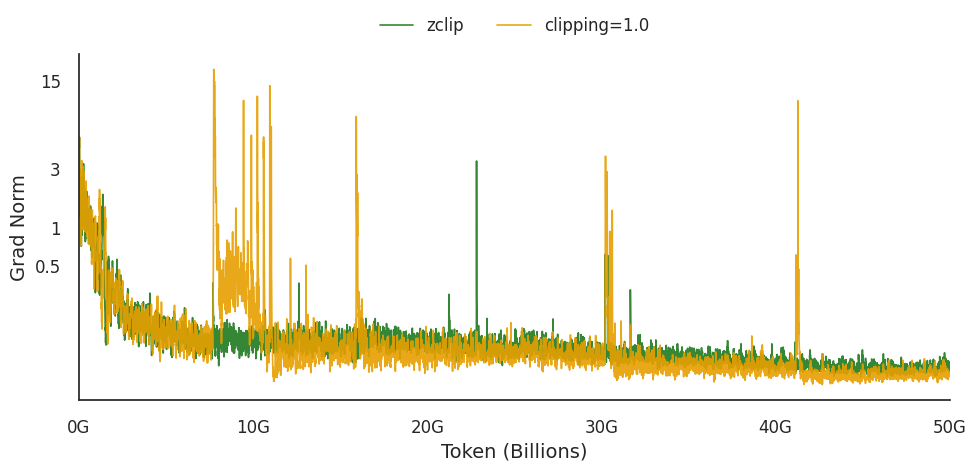}%
    }
    \subfloat[]{%
        \includegraphics[width=0.49\textwidth]{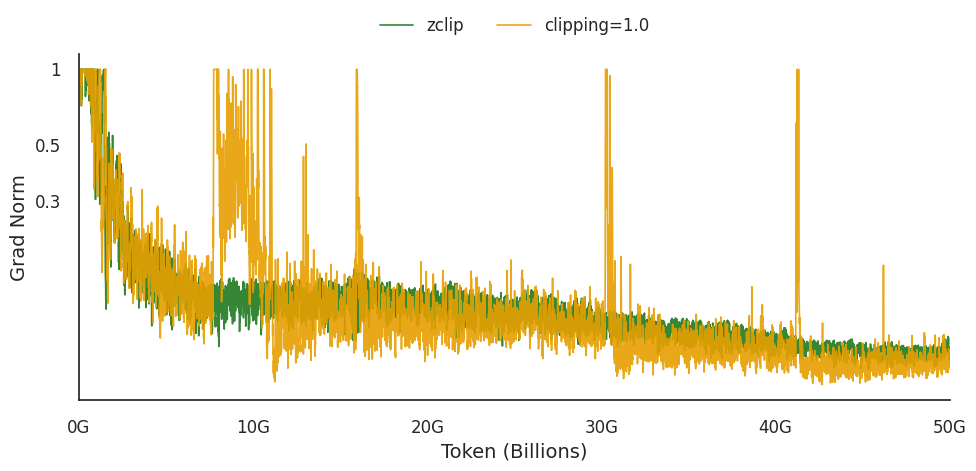}%
    }
    \caption{\textbf{Comparison of gradient norms for ZClip and fixed-threshold clipping with threshold value of 1.0, trained with a learning rate of 1e-3}. 
    (a) Before clipping: ZClip exhibits small, transient spikes early in training, while fixed-threshold clipping shows larger and more frequent deviations. 
    (b) After clipping: ZClip effectively suppresses these fluctuations, maintaining smooth and stable norms. 
    In contrast, fixed-threshold clipping fails to adapt to the evolving distribution, resulting in persistent post-clipping spikes and instability.}
  \label{fig:gradient_distribution}
\end{figure}

Figure~\ref{fig:gradient_distribution} provides a visual comparison of the gradient norm distribution as training progresses. 
Unlike fixed-threshold clipping, which suffered from wide variance and persistent post-clipping spikes due to its static threshold, ZClip produced a smoother and more controlled distribution. 
This demonstrated ZClip's ability to adaptively manage extreme outliers without suppressing smaller, informative updates, striking a balance that enhances both stability and performance. 

\section{Conclusion}

In this work, we introduced ZClip, an adaptive gradient clipping algorithm designed to address the limitations of fixed-threshold gradient norm clipping in training large-scale language models (LLMs). 
ZClip leverages z-scores for anomaly detection and tracks the evolution of mean and standard deviation using exponential moving averages (EMA). 
By dynamically adapting to the changing dynamics of gradient norms, ZClip effectively mitigates loss spikes and enhances training stability, making it particularly well suited for the challenges of modern LLM training.

In our experiments on 1B parameter LlaMA models, ZClip consistently outperformed both fixed-threshold clipping and percentile-based approaches. 
It expanded the space of feasible learning rates, enabling faster overall convergence.
In traditionally stable learning rate regimes, it mitigated spikes completely, and provided an uptick in downstream task performance.
In future work, we plan to present experiments across a wider set of architectures and larger model sizes, in particular at the 7B to 70B scale.
Furthermore, we are interested in evaluating ZClip in other traditionally noisy training scenarios such as reinforcement learning (RL) and multimodality.

\clearpage

\bibliographystyle{unsrt}

\section{Appendix}
\subsection{Model and Tokenizer Details}
\label{Model and Tokenizer Details}

\subsubsection*{Model Architecture}
\label{sec:llama_config}
Our model configuration is derived from a 1B parameter model class based on the LLaMA design as implemented in Hugging Face Transformers~\citep{wolf-etal-2020-transformers} version 4.47.1.
It consists of 16 Transformer decoder layers with a hidden size of 2048 and intermediate size of 5440. 
The model uses RMSNorm, SwiGLU activation, and supports rotary positional embeddings up to 2048 tokens.
The configuration parameters are summarized in Table~\ref{tab:model_config}. 

\begin{table}[H]
\centering
\begin{tabular}{|l|l|}
\hline
\textbf{Parameter} & \textbf{Value} \\
\hline
Hidden Size & 2048 \\
Intermediate Size (FFN) & 5440 \\
Number of Hidden Layers & 16 \\
Number of Attention Heads & 16 \\
Number of Key-Value Heads & 16 \\
Activation Function & SwiGLU \\
Normalization Type & RMSNorm \\
RMSNorm Epsilon & $1 \times 10^{-5}$ \\
Vocabulary Size & 65,536 \\
Max Position Embeddings & 2048 \\
\hline
\end{tabular}
\vspace{10pt}
\caption{\textbf{Model Configuration used for ZClip experiments.}}
\label{tab:model_config}
\end{table}

\noindent The use of SwiGLU activation improves gradient flow during training and complements the adaptive clipping properties of ZClip. RMSNorm provides a stable normalization baseline without introducing additional learnable parameters. 

\subsubsection*{Tokenizer}
\label{sec:tokenizer}
In our experiments, we used a custom tokenizer derived from the LLaMA 3 tokenizer. 
The tokenizer is truncated to include only the $65536 = 2^{16}$ most common tokens as computed on a random sample of 90\% FineWeb-Edu and 10\% Python-Edu. 
Other than that, it preserves key properties such as Unicode normalization, whitespace collapsing, and byte-pair encoding. 
This tokenizer was selected for its increased computational efficiency and compatibility with both code and natural language text.

\subsection{Parameterizing ZClip as a Percentile-based Approach}
\label{sec:zclip_percentile}
Rather than defining a threshold $z_{\text{thres}}$ in z-space, analogously to AutoClip one can define a maximum target percentile $p$ and derive the corresponding z-value \(z_p\) from the standard normal distribution for use in ZClip.
For example, for \(p=0.99\) one obtains \(z_{0.99} \approx 2.32635\).
In contrast to AutoClip which computes a clipping threshold by extracting an empirical percentile from the full history of gradient norms, the ZClip percentile approach still leverages exponentially weighted estimates of the gradient distribution to better account for the intrinsic shift in distribution as training progresses. 

\subsection{ZClip - ``Clipping to mean and max''}
\label{sec:clip-to-mean-max}
\begin{figure}[h]
    \centering
    \includegraphics[width=0.6\textwidth]{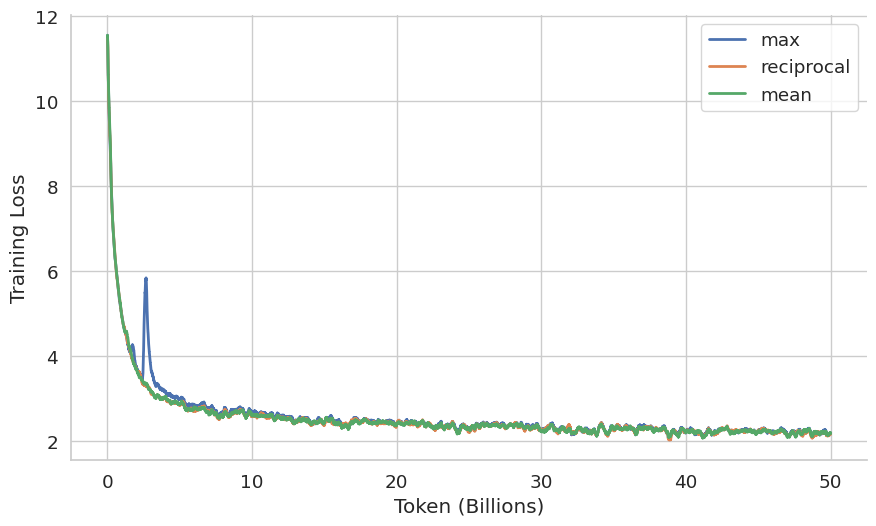}
    \caption{\textbf{Training loss for three ZClip variants—max, reciprocal, and mean.}
    The learning rate for all experiments is $1.0 \times 10^{-3}$.}
    \label{fig:clip_algo}
\end{figure}

For \emph{clipping to max}, i.e. $z_t^* = \mu_t + z_{\text{thres}} \cdot \sigma_t$, we still observed small spikes (see Figure~\ref{fig:clip_algo}).
\emph{clipping to mean}, on the other hand, is the most aggressive form of clipping. 
It eliminated spikes, but performed worse in downstream tasks (see Table~\ref{tab:clip_strat}).

\begin{table}[H]
\centering
\begin{tabular}{|l|c|c|c|c|}
\hline
\textbf{Clipping Strategy} & \textbf{Spike Count} & \textbf{Test Loss} & \textbf{HellaSwag (\%)} & 
\textbf{WinoGrande (\%)} \\
\hline
max      & 1 & 2.19 & 48.02  & 53.35  \\
mean     & \textbf{0} & \textbf{2.18}  & 48.90  & 54.22 \\
reciprocal     & \textbf{0} & \textbf{2.18}  & \textbf{49.30}  & \textbf{54.85} \\
\hline
\end{tabular}
\vspace{10pt}
\caption{\textbf{Downstream task performance on HellaSwag and Winogrande benchmarks compraing three ZClip variants—max, reciprocal, and mean.}
The learning rate for all experiments is $1.0 \times 10^{-3}$. 
The token budget was 50BT.}
\label{tab:clip_strat}
\end{table}

\subsection{Normality Assumption of Gradient Norms}
\label{appendix:normality}

ZClip relies on the assumption that gradient norms follow an approximately normal distribution over short temporal windows. 
This assumption underpins the use of z-scores for detecting statistical anomalies and guiding adaptive clipping thresholds.

\subsubsection{Empirical Validation}

\begin{figure}[htbp]
    \centering
    \subfloat[][]{%
        \includegraphics[width=0.6\textwidth]{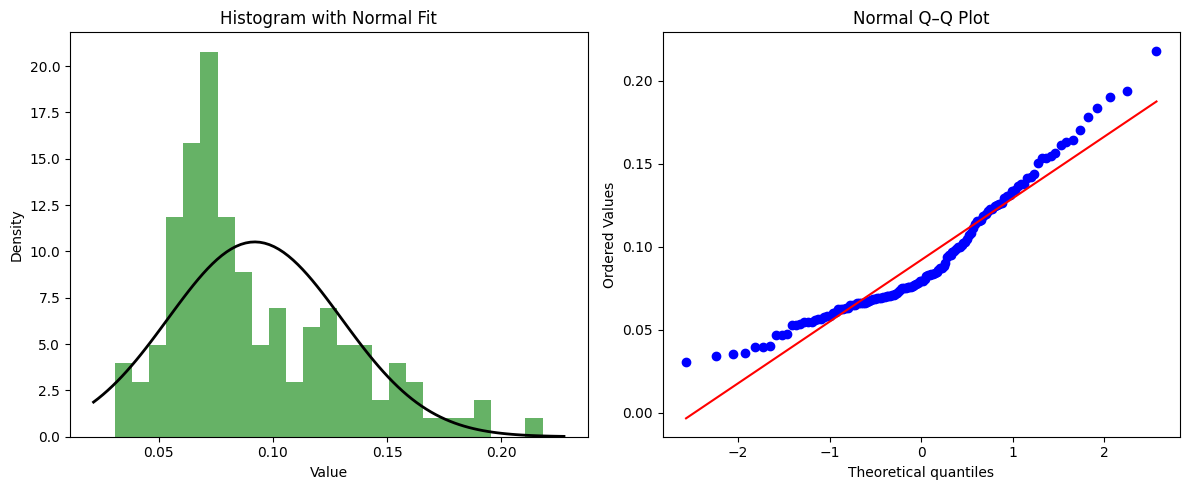}%
    }
    \hfill
    \subfloat[][]{%
        \includegraphics[width=0.6\textwidth]{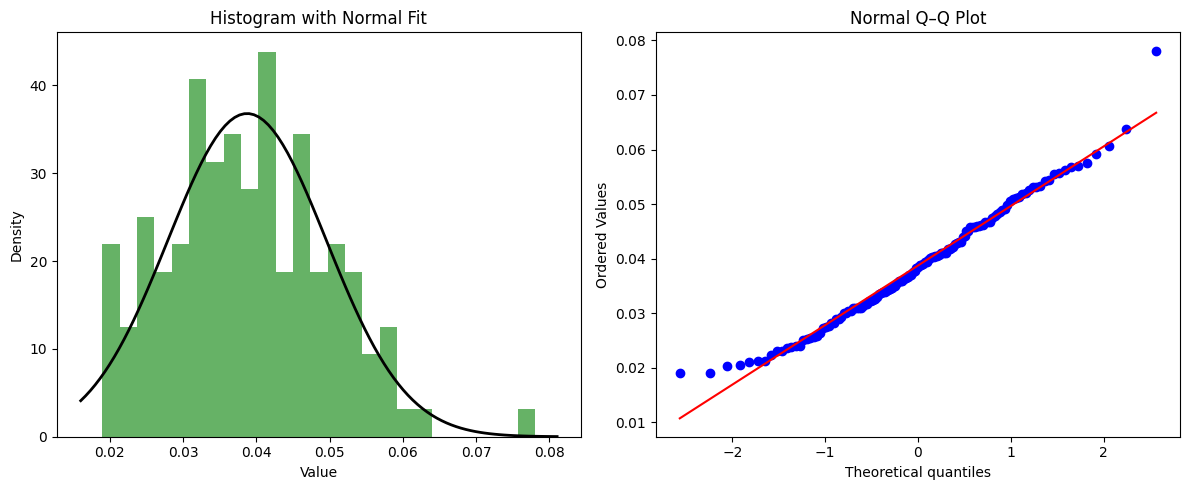}%
    }

    \caption{\textbf{Distribution of gradient norms in (a) Early training (steps 500–635) and (b) Mid training (steps 3050–3185).} The black curve shows the best-fit normal distribution. Early training exhibits mild skewness and heavier tails, while mid training displays improved symmetry and tighter variance.}   
    \label{fig:gradnorm_normality}
\end{figure}

To assess the validity of this assumption, we collected global gradient norms at various phases of training and fitted Gaussian curves to short-range sliding windows (e.g., 135 steps).
Figure~\ref{fig:gradnorm_normality} shows two representative examples.

\subsubsection{Interpretation and Relevance to ZClip}

In the early stages of training, we observe heavier tails and a mild right skew, reflecting initial instability and a learning rate ramp-up. In fact, when testing the distribution for training steps 500--635, the statistical tests did not support normality, indicating that the distribution did not yet conform to a normal curve. Despite this, the central mass of the distribution still closely approximated a normal curve, making z-score–based anomaly detection somewhat reliable even in imperfect settings. As training progresses and gradients stabilize, the distribution becomes tighter and more symmetric, reinforcing the validity of the normality assumption during mid- to late-stage training. To validate these assumptions, we conducted the Shapiro–Wilk test during training steps 3050--3185, which yielded a statistic of 0.9792 with a p-value of 0.0508.

We emphasize that ZClip does not require exact normality — it only requires that the distribution is sufficiently well behaved so that the mean and standard deviation can be used to derive a meaningful threshold. 
Although gradient norms are not perfectly Gaussian—particularly early in training—they seem to be sufficiently regular and unimodal to support robust, statistics-informed clipping using the z-score formulation.
In practice, the exponential moving average estimates of these statistics effectively smooth out noise and adapt to gradual distributional drift.

\subsection{Throughput Analysis}
\label{appendix:throughput}

\begin{figure}[H]
    \centering
    \includegraphics[width=0.55\textwidth]{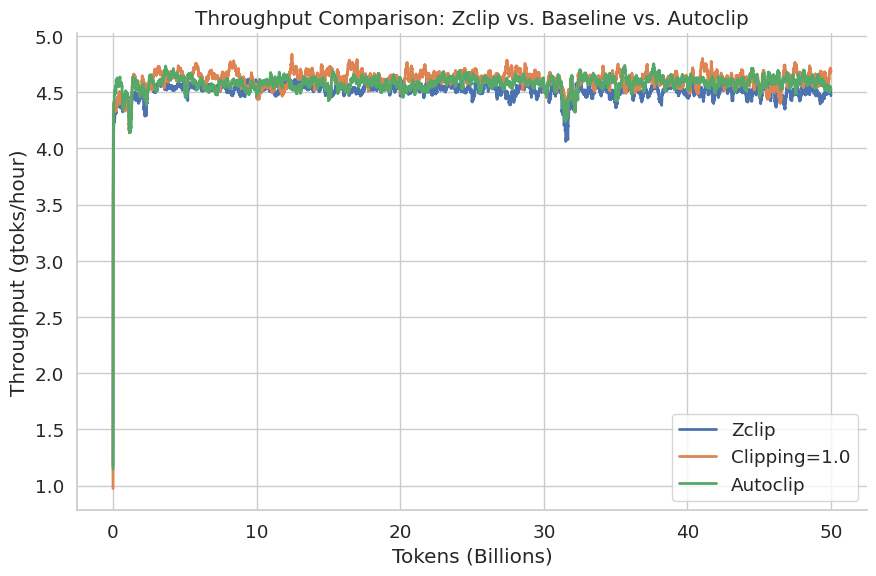}
    \caption{\textbf{Training throughput comparison between fixed-threshold gradient clipping, AutoClip, and ZClip methods.}}
    \label{fig:through}
\end{figure}

Throughput is a critical factor when training large language models. 
We evaluated the impact of enabling ZClip on training throughput in our experiments and observed that its lightweight, EMA-based computations introduced negligible overhead (see Figure~\ref{fig:through}).

\subsection{Hyperparameter Tuning}
\label{appendix:ablation}

\subsubsection{Z-Score Threshold}
The z-score threshold determines the degree of sensitivity towards anomalies in the gradient norms. 
Table~\ref{tab:zscore_threshold} presents the impact of varying the z-score threshold on training stability and convergence. 
Thresholds between 2.0 and 3.0 yielded the best overall trade-off, fully suppressing gradient spikes while preserving loss convergence and downstream performance. 
Lower thresholds introduce over-clipping, which can hinder learning, whereas higher thresholds fail to prevent occasional loss spikes. 

\begin{table}[H]
\centering
\begin{tabular}{|c|c|c|c|c|}
\toprule
\hline
\textbf{Z-Score Threshold} & \textbf{Train Loss} & \textbf{Spike Count} & \textbf{HellaSwag} & \textbf{Winogrande} \\
\hline
\midrule
1.5  & 2.182 & 0 & 48.70 & 52.23 \\
2.0  & \textbf{2.180} & 0 & 48.75 & 54.23 \\
\textbf{2.5}  & \textbf{2.180} & 0 & \textbf{49.30} & 54.85 \\
3.0  & 2.194 & 0 & 48.94 & 53.27 \\
3.5  & 2.202 & 1 & 48.30 & 53.67 \\
4.0  & 2.201 & 1 & 48.58 & \textbf{55.40} \\
\hline
\bottomrule
\end{tabular}
\vspace{10pt}
\caption{\textbf{Impact of Z-Score threshold on downstream task performance.} All models trained with a learning rate of $1.0 \times 10^{-3}$ and a token budget of 50B tokens.}
\label{tab:zscore_threshold}
\end{table}

\paragraph{Clipping Percentage Analysis}

\begin{figure}[H]
\centering
\includegraphics[width=0.6\textwidth]{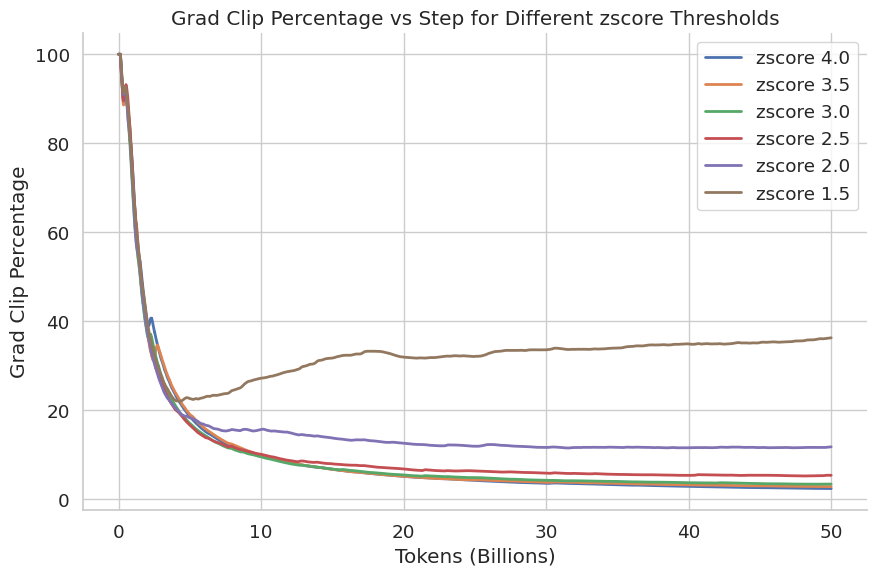}
\caption{\textbf{Clipping percentage vs.\ training step for different z-score thresholds}. Lower thresholds yield higher clipping percentages, indicating more aggressive clipping. The learning rate for all experiments is $1.0 \times 10^{-3}$. The token budget was 50BT.}
\label{fig:clipping_percentage}
\end{figure}

Figure~\ref{fig:clipping_percentage} illustrates how the proportion of clipped gradients evolves as training progresses for various z-score thresholds. 
When the threshold is lower (e.g., 1.5), a larger fraction of gradients are clipped, reflecting a more aggressive approach to suppressing potential outliers. 
Conversely, higher thresholds (e.g., 4.0) are more permissive and result in fewer clipped updates.

In our experiments, we tested thresholds ranging from 1.5 to 4.0 in increments of 0.5 and found that a threshold of 2.5 provides the most balanced trade-off. 
Specifically, at \(z_{\text{thres}} = 2.5\), the fraction of clipped gradients remains moderate, effectively removing harmful spikes without overly constraining gradients. 
This leads to more stable training and faster convergence compared to either excessively low or high thresholds.

\subsubsection{Alpha Value}
\begin{figure}[H]
\centering
\includegraphics[width=0.6\textwidth]{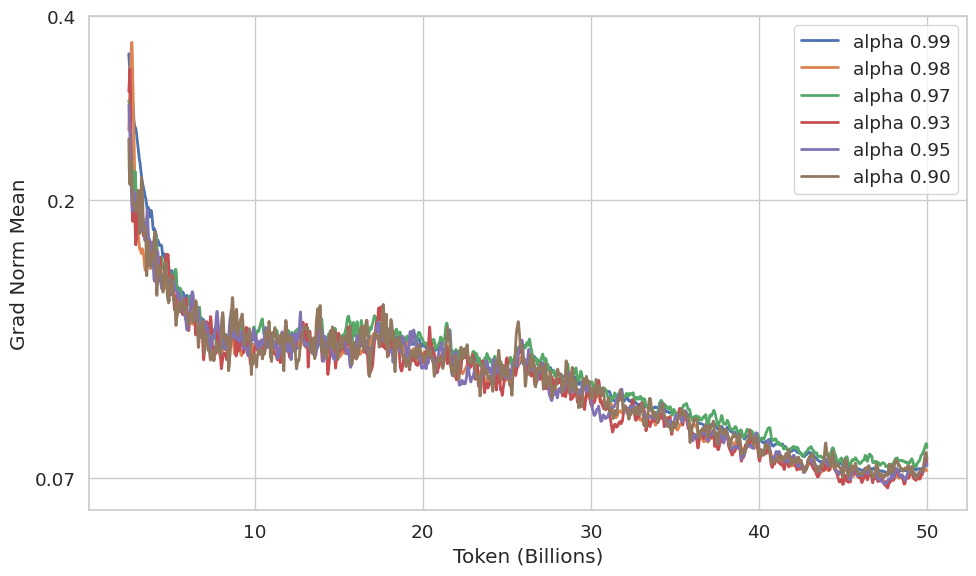}
\caption{\textbf{EMA-estimated mean for different \(\alpha\) values.} The learning rate for all experiments is $1.0 \times 10^{-3}$, and the threshold is $z_{\text{thres}} = 2.5$. The token budget was 50BT.}
\label{fig:ema_mean}
\end{figure}

\begin{figure}[H]
\centering
\includegraphics[width=0.6\textwidth]{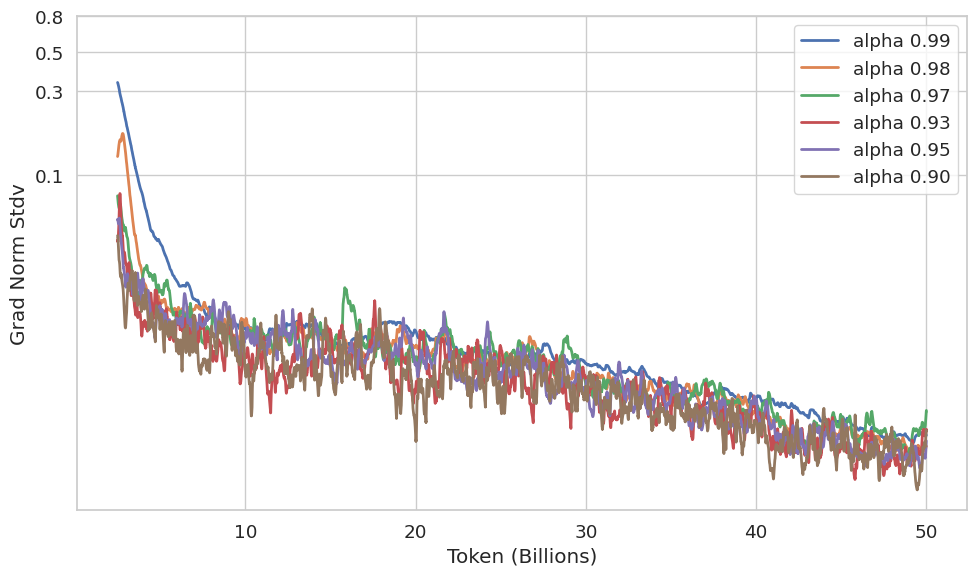}
\caption{\textbf{EMA-estimated standard deviation for different \(\alpha\) values.} The learning rate for all experiments is $1.0 \times 10^{-3}$, and the threshold is $z_{\text{thres}} = 2.5$. The token budget was 50BT.}
\label{fig:ema_std}
\end{figure}

We experimented with \(\alpha\) values ranging from \(0.9\) to \(0.99\) for ZClip's exponential moving average (EMA). 
A higher \(\alpha\) assigns more weight to historical gradients, leading to smoother updates but slower adaptation to changes in distribution over time. 
Conversely, lower \(\alpha\) values allow quicker adjustments but may introduce instability. 
Our results indicate that tuning \(\alpha\) appropriately helps balance stability and responsiveness in gradient clipping. 
Figure~\ref{fig:ema_mean} shows the effect of different \(\alpha\) values on the EMA-estimated mean, while Figure~\ref{fig:ema_std} illustrates the corresponding standard deviation estimates. Table~\ref{tab:alpha_abla} further shows that all configurations successfully suppressed gradient spikes, but the choice of alpha did affect downstream task performance. 
An intermediate setting of $\alpha = 0.97$ achieved the best trade-off.

\begin{table}[H]
\centering
\begin{tabular}{|c|c|c|c|c|}
\toprule
\hline
\textbf{Alpha} & \textbf{Train Loss} & \textbf{Spike Count} & \textbf{HellaSwag} & \textbf{Winogrande} \\
\hline
\midrule
0.90 & 2.190 & 0 & 48.04 & 54.01 \\
0.93 & 2.185 & 0 & 48.99 & 54.38 \\
0.95 & 2.186 & 0 & 48.91 & 53.98 \\
\textbf{0.97} & \textbf{2.180} & 0 & \textbf{49.30} & \textbf{54.85} \\
0.98 & 2.189 & 0 & 48.51 & 52.09 \\
0.99 & 2.189 & 0 & 48.45 & 53.74 \\
\bottomrule
\hline
\end{tabular}
\vspace{10pt}
\caption{\textbf{Impact of different values for $\alpha$ on stability and downstream performance.} All models trained with a learning rate of $1.0 \times 10^{-3}$ and a z-score threshold $z_{\text{thres}} = 2.5$. The token budget was 50B tokens.}
\label{tab:alpha_abla}
\end{table}

\end{document}